# The Replace Operator

**Lauri Karttunen**
Rank Xerox Research Centre
6, chemin de Maupertuis
F-38240 Meylan, France
`lauri.karttunen@xerox.fr`

**Abstract**

This paper introduces to the calculus of regular expressions a **replace** operator and defines a set of replacement expressions that concisely encode alternate variations of the operation. Replace expressions denote regular relations, defined in terms of other regular expression operators. The basic case is unconditional obligatory replacement. We develop several versions of conditional replacement that allow the operation to be constrained by context

## 0. Introduction

Linguistic descriptions in phonology, morphology, and syntax typically make use of an operation that replaces some symbol or sequence of symbols by another sequence or symbol. We consider here the replacement operation in the context of finite-state grammars.

Our purpose in this paper is twofold. One is to define replacement in a very general way, explicitly allowing replacement to be constrained by input and output contexts, as in two-level rules (Koskenniemi 1983), but without the restriction of only single-symbol replacements. The second objective is to define replacement within a general calculus of regular expressions so that replacements can be conveniently combined with other kinds of operations, such as composition and union, to form complex expressions.

Our replacement operators are close relatives of the rewrite-operator defined in Kaplan and Kay 1994, but they are not identical to it. We discuss their relationship in a section at the end of the paper.

### 0. 1. Simple regular expressions

The replacement operators are defined by means of regular expressions. Some of the operators we use to define them are specific to Xerox implementations of the finite-state calculus, but equivalent formulations could easily be found in other notations.

The table below describes the types of expressions and special symbols that are used to define the replacement operators.

[1]

| | |
|---|---|
| `(A)` | option (union of `A` with the empty string) |
| `~A` | complement (negation) |
| `\A` | term complement (any symbol other than `A`) |
| `$A` | contains (all strings containing at least one `A`) |
| `A*` | Kleene star |
| `A+` | Kleene plus |
| `A/B` | ignore (`A` interspersed with strings from `B`) |
| `A B` | concatenation |
| `A | B` | union |
| `A & B` | intersection |
| `A - B` | relative complement (minus) |
| `A .x. B` | crossproduct (Cartesian product) |
| `A .o. B` | composition |

Square brackets, `[]`, are used for grouping expressions. Thus `[A]` is equivalent to `A` while `(A)` is not. The order in the above table corresponds to the precedence of the operations. The prefix operators (`~`, `\`, and `$`) bind more tightly than the postfix operators (`*`, `+`, and `/`), which in turn rank above concatenation. Union, intersection, and relative complement are considered weaker than concatenation but stronger than crossproduct and composition. Operators sharing the same precedence are interpreted left-to-right. Our new replacement operator goes in a class between the Boolean operators and composition. Taking advantage of all these conventions, the fully bracketed expression

[2]

```
[[[~[a]]* [[b]/x]] | c] .x. d ;
```

can be rewritten more concisely as

[3]

```
~a* b/x | c .x. d    ;
```

Expressions that contain the crossproduct (`.x.`) or the composition (`.o.`) operator describe regular relations rather than regular languages. A regular relation is a mapping from one regular language to another one. Regular languages correspond to simple finite-state automata; regular relations are modeled by finite-state transducers. In the relation `A .x. B`, we call the first member, `A`, the **upper** language and the second member, `B`, the **lower** language.

To make the notation less cumbersome, we systematically ignore the distinction between the language `A` and the identity relation that maps every string of `A` to itself. Correspondingly, a simple automaton may be thought of as representing a language or as a transducer for its identity relation. For the sake of convenience, we also equate a language consisting of a single string with the string itself. Thus the expression `abc` may denote, depending on the context, (i) the string `abc`, (ii) the language consisting of the string `abc`, and (iii) the identity relation on that language.

We recognize two kinds of symbols: simple symbols (`a`, `b`, `c`, etc.) and fst pairs (`a:b`, `y:z`, etc.). An fst pair `a:b` can be thought of as the crossproduct of `a` and `b`, the minimal relation consisting of `a` (the upper symbol) and `b` (the lower symbol). Because we regard the identity relation on `A` as equivalent to `A`, we write `a:a` as just `a`. There are two special symbols

[4]

- `0`    epsilon (the empty string).
- `?`    any symbol in the known alphabet and its extensions.

The escape character, `%`, allows letters that have a special meaning in the calculus to be used as ordinary symbols. Thus `%&` denotes a literal ampersand as opposed to `&`, the intersection operator; `%0` is the ordinary zero symbol.

The following simple expressions appear frequently in our formulas:

[5]

- `[]`    the empty string language.
- `~$[]`    the null set.
- `?*`    the universal ("sigma-star") language: all possible strings of any length including the empty string.

### 1. Unconditional replacement

To the regular-expression language described above, we add the new replacement operator. The unconditional replacement of UPPER by LOWER is written

[6]

```
UPPER -> LOWER
```

Here `UPPER` and `LOWER` are any regular expressions that describe simple regular languages. We define this replacement expression as

[7]

```
[ NO_UPPER   [UPPER .x. LOWER] ]*
NO_UPPER ;
```

where `NO_UPPER` abbreviates `~$[UPPER - []]`. The definition describes a regular relation whose members contain any number (including zero) of iterations of `[UPPER .x. LOWER]`, possibly alternating with strings not containing `UPPER` that are mapped to themselves.

### 1.1. Examples

We illustrate the meaning of the replacement operator with a few simple examples. The regular expression

[8]

```
a b | c -> x ;
(same as [[a b] | c] -> x)
```

describes a relation consisting of an infinite set of pairs such as

[9]

```
a b a c a
x   a x a
```

where all occurrences of `ab` and `c` are mapped to `x` interspersed with unchanging pairings. It also includes all possible pairs like

[10]

```
x a x a
x a x a
```

that do not contain either `ab` or `c` anywhere.

Figure 1 shows the state diagram of a transducer that encodes this relation. The transducer consists of states and arcs that indicate a transition from state to state over a given pair of symbols. For convenience we represent identity pairs by a single symbol; for example, we write `a:a` as `a`. The symbol `?` represents here the identity pairs of symbols that are not explicitly present in the network. In this case, `?` stands for any identity pair other than `a:a`, `b:b`, `c:c`, and `x:x`. Transitions that differ only with respect to the label are collapsed into a single multiply labelled arc. The state labeled `0` is the start state. Final states are distinguished by a double circle.

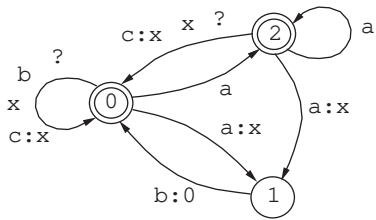

Figure 1:    a b | c -> x

Every pair of strings in the relation corresponds to a path from the initial `0` state of the transducer to a final state. The `abaca` to `xaxa` path is `0-1-0-2-0-2`, where the `2-0` transition is over a `c:x` arc.

In case a given input string matches the replacement relation in two ways, two outputs are produced. For example,

[11]
```
a b | b c -> x   ;
```

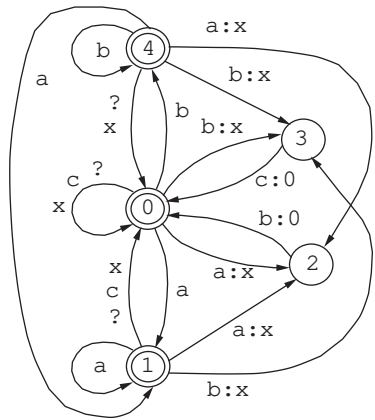

Figure 2: a b | b c -> x

maps `abc` to both `ax` and `xc`:

[12]
```
a b c      ,    a b c
a x                x c
```

The corresponding transducer paths in Figure 2 are `0-1-3-0` and `0-2-0-0`, where the last `0-0` transition is over a `c` arc.

If this ambiguity is not desirable , we may write two replacement expressions and compose them to indicate which replacement should be preferred if a choice has to be made. For example, if the `ab` match should have precedence, we write

[13]
```
a b -> x
  .o.
b c -> x   ;
```

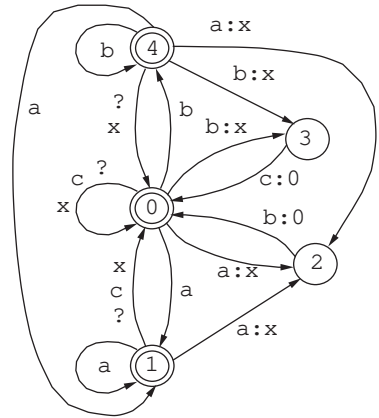

Figure 3: a b -> x .o. b c -> x

This composite relation produces the same output as the previous one except for strings like `abc` where it unambiguously makes only the first replacement, giving `xc` as the output. The `abc` to `xc` path in Figure 3 is `0-2-0-0`.

### 1.2. Special cases

Let us illustrate the meaning of the replacement operator by considering what our definition implies in a few special cases.

If UPPER is the empty set, as in

[14]
```
[] -> a | b      ;
```

the expression compiles to a transducer that freely inserts `a`s and `b`s in the input string.

If UPPER describes the null set, as in,

[15]
```
~$[] -> a | b    ;
```

the `LOWER` part is irrelevant because there is no replacement. This expression is a description of the sigma-star language.

If `LOWER` describes the empty set, replacement becomes deletion. For example,

[16]
```
  a | b -> []       ;
```

removes all `a`s and `b`s from the input.

If `LOWER` describes the null set, as in all strings containing UPPER, here `a` or `b`, are excluded from the upper side language. Everything else is mapped to itself. An equivalent expression is `~$[a | b]`.

[17]
```
  a | b -> ~$[]    ;
```

### 1.3. Inverse replacement

The inverse replacement operator.

[18]
```
  UPPER <- LOWER
```

is defined as the inverse of the relation `LOWER -> UPPER`.

### 1.4. Optional replacement

An optional version of unconditional replacement is derived simply by replacing `NO_UPPER` in [7] with `?*`, the sigma-star language.

[19]
```
  UPPER (->) LOWER
```

is defined as

[20]
```
  [ ?* [UPPER .x. LOWER] ]* ?*    ;
```

The optional replacement relation maps `UPPER` to both `LOWER` and itself. The optional version of `<-` is defined in the same way.

### 2. Conditional replacement

We now extend the notion of simple replacement by allowing the operation to be constrained by a left and a right context. A conditional replacement expression has four components: `UPPER`, `LOWER`, `LEFT`, and `RIGHT`. They must all be regular expressions that describe a simple language. We write the replacement part `UPPER -> LOWER`, as before, and the context part as `LEFT _ RIGHT`, where the underscore indicates where the replacement takes place.

In addition, we need a separator between the replacement and the context part. We use four alternate separators, `||`, `//`, `\\` and `\/`, which gives rise to four types of conditional replacement expressions:

[21]
(1) Upward-oriented:
`UPPER -> LOWER || LEFT _ RIGHT ;`

(2) Right-oriented:
`UPPER -> LOWER // LEFT _ RIGHT ;`

(3) Left-oriented:
`UPPER -> LOWER \\ LEFT _ RIGHT ;`

(4) Downward-oriented:
`UPPER -> LOWER \/ LEFT _ RIGHT ;`

All four kinds of replacement expressions describe a relation that maps `UPPER` to `LOWER` between `LEFT` and `RIGHT` leaving everything else unchanged. The difference is in the interpretation of "between `LEFT` and `RIGHT`."

### 2.1. Overview: divide and conquer

We define `UPPER -> LOWER || LEFT _ RIGHT` and the other versions of conditional replacement in terms of expressions that are already in our regular expression language, including the unconditional version just defined. Our general intention is to make the conditional replacement behave exactly like unconditional replacement except that the operation does not take place unless the specified context is present.

This may seem a simple matter but it is not, as Kaplan and Kay 1994 show. There are several sources of complexity. One is that the part that is being replaced may at the same time serve as the context of another adjacent replacement. Another complication is the fact just mentioned: there are several ways to constrain a replacement by a context.

We solve both problems using a technique that was originally invented for the implementation of phonological rewrite rules (Kaplan and Kay 1981, 1994) and later adapted for two-level rules (Kaplan, Karttunen, Koskenniemi 1987; Karttunen and

Beesley 1992). The strategy is first to decompose the complex relation into a set of relatively simple components, define the components independently of one another, and then define the whole operation as a composition of these auxiliary relations.

We need six intermediate relations, to be defined shortly:

[22]

(1) InsertBrackets
(2) ConstrainBrackets
(3) LeftContext
(4) RightContext
(5) Replace
(6) RemoveBrackets

Relations (1), (5), and (6) involve the unconditional replacement operator defined in the previous section.

Two auxiliary symbols, < and >, are introduced in (1) and (6). The left bracket, <, indicates the end of a left context. The right bracket, >, marks the beginning of a complete right context. The distribution of the auxiliary brackets is controlled by (2), (3), and (4). The relations (1) and (6) that introduce the brackets internal to the composition at the same time remove them from the result.

**2.2. Basic definition**

The full specification of the six component relations is given below. Here UPPER, LOWER, LEFT, and RIGHT are placeholders for regular expressions of any complexity.

In each case we give a regular expression that precisely defines the component followed by an English sentence describing the same language or relation. In our regular expression language, we have to prefix the auxiliary context markers with the escape symbol % to distinguish them from other uses of < and >.

[23]

(1) InsertBrackets

```
[] <- %< | %> ;
```

*The relation that eliminates from the upper side language all context markers that appear on the lower side.*

[24]

(2) ConstrainBrackets

```
~$[%< %>] ;
```

*The language consisting of strings that do not contain <> anywhere.*

[25]

(3) LeftContext

```
~[~[...LEFT]   [<...]] &
~[ [...LEFT] ~[<...]] ;
```

*The language in which any instance of < is immediately preceded by LEFT, and every LEFT is immediately followed by <, ignoring irrelevant brackets.*

Here [...LEFT] is an abbreviation for [[?* LEFT/[%<|%>]] - [?* %<]], that is, any string ending in LEFT, ignoring all brackets except for a final <. Similarly, [%<...] stands for [%</%> ?*], any string beginning with <, ignoring the other bracket.

[26]

(4) RightContext

```
~[ [...>] ~[RIGHT...]] &
~[~[...>]   [RIGHT...]] ;
```

*The language in which any instance of > is immediately followed by RIGHT, and any RIGHT is immediately preceded by >, ignoring irrelevant brackets.*

Here [...>] abbreviates [?* %>/%<], and RIGHT... stands for [RIGHT/[%<|%>] - [%> ?*]], that is, any string beginning with RIGHT, ignoring all brackets except for an initial >.

[27]

(5) Replace

```
%< UPPER/[%<|%>] %>
      ->
%< LOWER/[%<|%>] %> ;
```

*The unconditional replacement of <UPPER> by <LOWER>, ignoring irrelevant brackets.*

The redundant brackets on the lower side are important for the other versions of the operation.

[28]

(6) RemoveBrackets

```
%< | %> -> [] ;
```

*The relation that maps the strings of the upper language to the same strings without any context markers.*

The upper side brackets are eliminated by the inverse replacement defined in (1).

### 2.3. Four ways of using contexts

The complete definition of the first version of conditional replacement is the composition of these six relations:

[29]
```
UPPER -> LOWER || LEFT _ RIGHT ;

  InsertBrackets
       .o.
 ConstrainBrackets
       .o.
    LeftContext
       .o.
   RightContext
       .o.
     Replace
       .o.
  RemoveBrackets ;
```

The composition with the left and right context constraints prior to the replacement means that any instance of UPPER that is subject to replacement is surrounded by the proper context on the upper side. Within this region, replacement operates just as it does in the unconditional case.

Three other versions of conditional replacement can be defined by applying one, or the other, or both context constraints on the lower side of the relation. It is done by varying the order of the three middle relations in the composition. In the right-oriented version (//), the left context is checked on the lower side of replacement:

[30]
```
UPPER -> LOWER // LEFT _ RIGHT ;

     ...
  RightContext
      .o.
    Replace
      .o.
   LeftContext
     ...
```

The left-oriented version applies the constraints in the opposite order:

[31]
```
UPPER -> LOWER \\ LEFT _ RIGHT ;

     ...
   LeftContext
      .o.
    Replace
      .o.
  RightContext
     ...
```

The first three versions roughly correspond to the three alternative interpretations of phonological rewrite rules discussed in Kaplan and Kay 1994. The upward-oriented version corresponds to simultaneous rule application; the right- and left-oriented versions can model rightward or leftward iterating processes, such as vowel harmony and assimilation.

The fourth logical possibility is that the replacement operation is constrained by the lower context.

[32]
```
UPPER -> LOWER \/ LEFT _ RIGHT ;

     ...
    Replace
      .o.
   LeftContext
      .o.
  RightContext
     ...
```

When the component relations are composed together in this manner, UPPER gets mapped to LOWER just in case it ends up between LEFT and RIGHT in the output string.

### 2.4. Examples

Let us illustrate the consequences of these definitions with a few examples. We consider four versions of the same replacement expression, starting with the upward-oriented version

[33]
```
a b -> x || a b _ a ;
```

applied to the string ababab. The resulting relation is

[34]
```
a b a b a b a
a b  x   x  a
```

The second and the third occurrence of ab are replaced by x here because they are between ab and x on the upper side language of the relation. A transducer for the relation is shown in Figure 4.

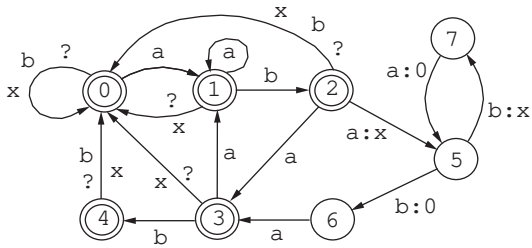

Figure 4: a b -> x || a b _ a

The path through the network that maps `abababa` to `abxxa` is `0-1-2-5-7-5-6-3`.

The right-oriented version,

[35]
```
a b -> x // a b _ a;
```

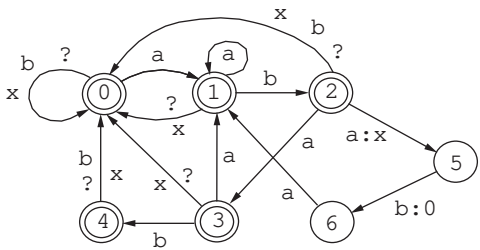

Figure 5: a b -> x // a b _ a

gives us a different result:

[36]
```
a b a b a b a
a b   x   a b a
```

following the path `0-1-2-5-6-1-2-3`. The last occurrence of ab must remain unchanged because it does not have the required left context on the lower side.

The left-oriented version of the rule shows the opposite behavior because it constrains the left context on the upper side of the replacement relation and the right context on the lower side.

[37]
```
a b -> x \\ a b _ a ;
```

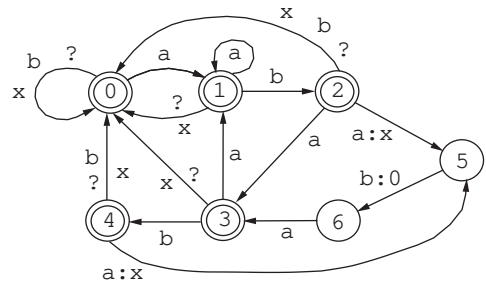

Figure 6: a b -> x \\ a b _ a

With `abababa` composed on the upper side, it yields

[38]
```
a b a b a b a
a b a b   x   a
```

by the path `0-1-2-3-4-5-6-3`.

The first two occurrences of ab remain unchanged because neither one has the proper right context on the lower side to be replaced by x.

Finally, the downward-oriented fourth version:

[39]
```
a b -> x \/ a b _ a ;
```

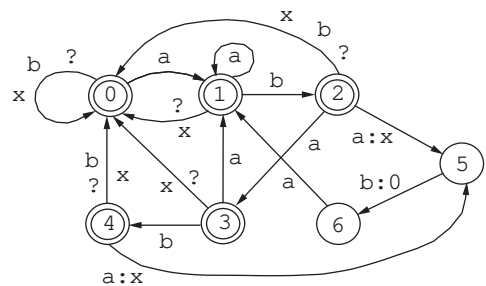

Figure 7: a b -> x \/ a b _ a

This time, surprisingly, we get two outputs from the same input:

[40]
```
a b a b a b a  ,  a b a b a b a
a b   x   a b a      a b a b   x   a
```

Path `0-1-2-5-6-1-2-3` yields `abxaba`, path `0-1-2-3-4-5-6-1` gives us `ababxa`

It is easy to see that if the constraint for the replacement pertains to the lower side, then in this case it can be satisfied in two ways.

## 3. Comparisons

### 3.1. Phonological rewrite rules

Our definition of replacement is in its technical aspects very closely related to the way phonological rewrite-rules are defined in Kaplan and Kay 1994 but there are important differences. The initial motivation in their original 1981 presentation was to model a left-to-right deterministic process of rule application. In the course of exploring the issues, Kaplan and Kay developed a more abstract notion of rewrite rules, which we exploit here, but their 1994 paper retains the procedural point of view.

Our paper has a very different starting point. The basic case for us is unconditional obligatory replacement, defined in a purely relational way without any consideration of how it might be applied. By starting with obligatory replacement, we can easily define an optional version of the operator. For Kaplan and Kay, the primary notion is optional rewriting. It is quite cumbersome for them to provide an obligatory version. The results are not equivalent.

Although people may agree, in the case of simple phonological rewrite rules, what the outcome of a deterministic rewrite operation should be, it is not clear that this is the case for replacement expressions that involve arbitrary regular languages. For that reason, we prefer to define the replacement operator in relational terms without relying on an uncertain intuition about a particular procedure.

### 3.2. Two-level rules

Our definition of replacement also has a close connection to two-level rules. A two-level rule always specifies whether a context element belongs to the input (= lexical) or the output (= surface) context of the rule. The two-level model also shares our pure relational view of replacement as it is not concerned about the application procedure. But the two-level formalism is only defined for symbol-to-symbol replacements.

## 4. Conclusion

The goal of this paper has been to introduce to the calculus of regular expressions a replace operator, ->, with a set of associated replacement expressions that concisely encode alternate variations of the operation.

We defined unconditional and conditional replacement, taking the unconditional obligatory replacement as the basic case. We provide a simple declarative definition for it, easily expressed in terms of the other regular expression operators, and extend it to the conditional case providing four ways to constrain replacement by a context.

These definitions have already been implemented. The figures in this paper correspond exactly to the output of the regular expression compiler in the Xerox finite-state calculus.

## Acknowledgments

This work is based on many years of productive collaboration with Ronald M. Kaplan and Martin Kay. I am particularly indebted to Kaplan for writing a very helpful critique, even though he strongly prefers the approach of Kaplan and Kay 1994. Special thanks are also due to Kenneth R. Beesley for help on the definitions of the replace operators and for expert editorial advice. I am grateful to Jean-Pierre Chanod, Makoto Kanazawa, André Kempe, Pasi Tapanainen, and Annie Zaenen for helping to correct many technical and rhetorical weaknesses of the initial draft .